\documentclass[runningheads,a4paper]{llncs}

\usepackage{amssymb}
\usepackage{graphicx}

\usepackage{url}

\newcommand{\keywords}[1]{\par\addvspace\baselineskip
\noindent\keywordname\enspace\ignorespaces#1}

\usepackage{bbm}
\usepackage{graphicx}
\usepackage{booktabs}
\usepackage{multirow}

\usepackage{amsmath}

\usepackage[tight,footnotesize]{subfigure}

\usepackage{url}
\usepackage{amssymb}
\usepackage{times}
\usepackage{algorithm}
\usepackage{algorithmic}
\usepackage{enumerate}

\usepackage{color}
\usepackage{float}
\usepackage{dblfloatfix}
\usepackage{fixltx2e}

\newcommand{\xbf}{\ensuremath{\mathbf{x}}}
\newcommand{\Dcal}{\ensuremath{\mathcal{D}}}

\newcommand{\Qcal}{\ensuremath{\mathcal{Q}}}
\newcommand{\Wcal}{\ensuremath{\mathcal{Q}}}
\newcommand{\Qbf}{\ensuremath{\mathbf{Q}}}

\newcommand{\Mbf}{\ensuremath{\mathbf{M}}}
\newcommand{\mbf}{\ensuremath{\mathbf{m}}}
\newcommand{\Abf}{\ensuremath{\mathbf{A}}}
\newcommand{\Hcal}{\ensuremath{\mathcal{H}}}
\newcommand{\R}{\ensuremath{\mathbb{R}}}

\newcommand{\PR}{\operatorname{\mathbb{P}}}
\newcommand{\BQ}{B_{\Qcal}}
\newcommand{\posterior}{\ensuremath{\mathcal{Q}}}
\newcommand{\RP}{R_D}

\newcommand{\Var}{\operatorname{\mathbf{Var}}}
\newcommand{\sign}{\operatorname{sign}}
\newcommand{\argmin}{\operatorname{argmin}}
\newcommand{\argmax}{\operatorname{argmax}}

\newcommand{\x}{\times}

\newcommand{\dive}[2]{
  \operatorname{diff}_{#1}^{#2}
}

\newcommand{\esp}[1]{
    \underset{#1}{\mathbb{E}}\
}
\newcommand{\espdevant}[1]{
    \mathbb{E}_{#1}\,
}

\makeatletter
\def\captionof#1#2{{\def\@captype{#1}#2}}
\makeatother

\begin{document}

\mainmatter  

\title{Majority Vote of Diverse Classifiers for Late Fusion}

\titlerunning{Majority Vote of Diverse Classifiers for Late Fusion}

%
%

\author{Emilie Morvant\inst{1}
\and Amaury Habrard\inst{2} \and St{\'e}phane Ayache\inst{3}}
\authorrunning{Weighted Majority Vote of Diverse Classifiers for Late Fusion}
\institute{Institute of Science and Technology (IST) Austria,  A-3400 Klosterneuburg, Austria
\and
Universit{\'e} de Saint-Etienne, CNRS, LaHC,  UMR 5516,  F-42000 St-Etienne, France
\and
Aix-Marseille Universit{\'e}, CNRS, LIF UMR 7279, 
F-13000, Marseille, France
}

\maketitle

\begin{abstract}
In the past few years, a lot of attention has been devoted to multimedia indexing by fusing multimodal informations.
Two kinds of fusion schemes are generally considered: The {\it early fusion} and the {\it late fusion}.
We focus on late classifier fusion, where one combines the scores of each modality at the decision level.
To tackle this problem, we investigate a recent and elegant well-founded quadratic program named MinCq coming from the machine learning PAC-Bayesian theory.
MinCq looks for the weighted combination, over a set of real-valued functions seen as voters, leading to the lowest misclassification rate, while maximizing the voters' diversity.
We propose an extension of MinCq tailored to multimedia indexing.
Our method is based on an order-preserving pairwise loss adapted to ranking that allows us to improve Mean Averaged Precision measure while taking into account the diversity of the voters that we want to fuse.
We provide evidence that this method is naturally adapted to late fusion procedures and confirm the good behavior of our approach on the challenging PASCAL VOC'07 benchmark.
\end{abstract}

\section{Introduction}

\begin{figure}[t]
\label{fig:fusion}
\centering
\includegraphics[width=0.55\linewidth]{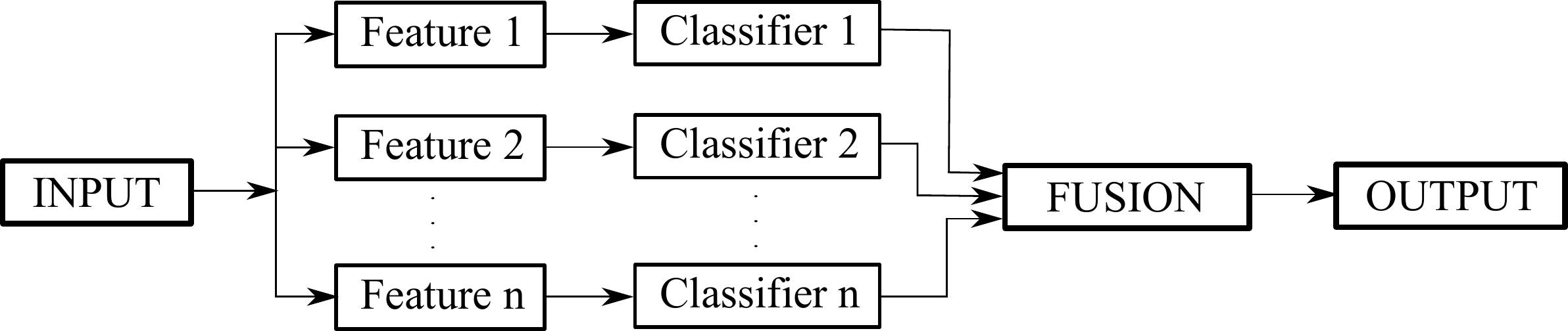}
\caption{Classical late classifier fusion scheme.}
\end{figure}

Combining multimodal information is an important issue in pattern recognition. 
Indeed, the fusion of multimodal inputs can bring
complementary information from various  sources, useful for improving
the quality of the final decision.
In this paper, we focus on multimodal fusion for image analysis in multimedia systems (see \cite{AtreyHEK10} for a survey). 
The different modalities correspond generally to a relevant set of
features that can be grouped into views.
Once these features have been extracted, another step consists in using
machine learning methods in order to build voters---or classifiers---able to
discriminate a given concept. 
In this context, two main schemes are generally considered \cite{Early-Late-ACMMultimedia05}. 
On the one hand, in the {\it early fusion} approach, all the available
features are merged into one feature vector before the learning and
classification steps. This can be seen as a unimodal classification. 
However, this kind of approach has to deal with many heterogeneous 
features which are sometimes hard to combine. 
On the other hand, the {\it late fusion}\footnote{The late fusion is sometimes called
multimodal classification or classifier fusion.}  
 works at the decision level by
combining the prediction scores available for each
modality (see Fig.~\ref{fig:fusion}).  
Even if late fusion  may not always outperform 
early fusion\footnote{For example, when one modality provides
significantly better results than others.}, it tends to give 
better results in multimedia analysis~\cite{Early-Late-ACMMultimedia05}. 
Several methods based on a fixed decision rule have been proposed for combining classifiers such as {\tt max}, {\tt min}, {\tt sum}, etc \cite{kittler98b}.
Other approaches, often referred to as {\it stacking} \cite{Wolpert92}, need of an extra learning step.

\indent In this paper, we address the problem of {\it late fusion} with stacking. Let $h_i$ be the function that gives the score associated with the $i^{th}$ modality for any instance $\xbf$. 
A classical method consists in looking for a weighted linear combination of the  scores seen as a majority vote and defined by: $ H(\xbf)\!\!=\!\! \sum_{i=1}^n q_i h_i(\xbf),$ where $q_i$ is the weight associated with $h_i$. 
This approach is widely used because of its robustness, simplicity and scalability due to small computational costs \cite{AtreyHEK10}. It is also more appropriate when there exist dependencies between the views through the classifiers \cite{WuCCS04,ma2013linear}. 
The objective is then to find an optimal combination of the classifiers' scores by taking into account these dependencies. One solution is to use machine learning methods to assess automatically the weights \cite{Kuncheva,Dietterich00,re2012ensemble,sun2013survey}.
Indeed, from a theoretical machine learning standpoint: considering a set of classifiers with a high diversity is a desirable property \cite{Dietterich00}. 
One illustration  is given by the algorithm AdaBoost \cite{FreundS96} that  weights {\it weak classifiers} according to different distributions of the training data, introducing some diversity. However,  AdaBoost degrades the fusion performance when combining strong classifiers \cite{Wickram01}.
\\\indent To tackle the late fusion by taking into account the diversity between score functions of strong classifiers, we propose a new framework based on a recent machine learning algorithm called MinCq \cite{MinCQ}. 
MinCq is expressed as a quadratic program for learning a weighted majority vote over real-valued functions called voters (such as score functions of classifiers). The algorithm is based on the minimization of a generalization bound that takes into account both the risk of committing an error and  the diversity of the voters, offering strong theoretical guarantees on the learned majority vote. In this article, our aim is to show the usefulness of MinCq-based methods for classifier fusion. We provide evidence that they are  able to find good linear weightings, and also performing non-linear combination with an extra kernel layer over the scores.
Moreover, since in multimedia retrieval, the performance measure is related to the rank of positive examples, we extend MinCq to optimize the Mean Average Precision. We base this extension on an additional order-preserving loss for verifying ranking pairwise constraints.

The paper is organized as follows. The framework of MinCq is introduced in Section~\ref{sec:mincq}. Our extension for late classifier fusion is presented in Section \ref{sec:mincqpw} and it is evaluated on an image annotation task in Section \ref{sec:expe}. We conclude in Section \ref{sec:conclu}. 

\section{MinCq: A Quadratic Program for Majority Votes}
\label{sec:mincq}
We start from the presentation of MinCq  \cite{MinCQ}, a quadratic program for learning a weighted majority vote of real-valued functions for binary classification.
Note that this method is based on the machine learning PAC-Bayesian theory, first introduced in \cite{Mcallester99b}. 

We consider binary classification tasks over a {\it feature space} $X\!\subseteq\!\R^d$ of dimension $d$.
The {\it label space} (or output space) is  $Y\!=\!\{-1,1\}$.
The training sample of size $m$ is 
$S\!=\!\{(\xbf_i,y_i)\}_{i=1}^{m}$ where each example $(\xbf_i,y_i)$ is drawn {\it i.i.d.} from a fixed---but unknown---probability distribution $\Dcal$ defined over $X\!\x\! Y$.
We consider a set of $n$ real-valued voters $\Hcal$, such that: $\forall h_i \in \Hcal,\ h_i\!:\!X\!\mapsto\!\R$.
Given a voter $h_i\in\Hcal$, the predicted label of $\xbf\!\in\! X$ is given by $\sign[h_i(\xbf)]$, where $\sign[a]\!=\!1$ if $a\!\geq\! 0$ and $-1$ otherwise. 
Then, the learner aims at choosing the weights $q_i$, leading to the {\it $\Qcal$-weighted majority vote} $B_{\Qcal}$ with the lowest risk. In the specific setting of MinCq\footnote{In PAC-Bayes these weights are modeled by a distribution $\Qcal$ over $\Hcal$ s.t. $\forall h_i\in \Hcal,\ q_i\!=\!\Qcal(h_i)$.}, $B_{\Qcal}$ is defined by,
\centerline{
$\displaystyle B_{\Qcal}(\xbf) = \sign\left[H_{\Qcal}(\xbf)\right]\textrm{, with }H_{\Qcal}(\xbf)=\sum_{i=1}^{n}q_i h_i(\xbf),
$}\\
where $\forall i\in\{1,\dots,n\},\ \sum_{i=1}^n \left|q_i\right| = 1$ and $-1\leq q_i\leq 1$. 
Its true risk $R_{\Dcal}(B_{\Qcal})$ is defined as the probability that $\BQ$  misclassifies an example drawn according to $\Dcal$,\\[1mm]
\centerline{$\displaystyle R_{\Dcal}(B_{\Qcal}) = \PR_{(\xbf,y)\sim \mathcal{D}} \left(B_{\Qcal}(\xbf)\ne y\right).$}\\[1mm]
\indent The core of MinCq  is  the minimization of the empirical version of a bound---the $C$-Bound \cite{Lacasse07,MinCQ}---over  $R_{\Dcal}(B_{\Qcal})$. 
The $C$-Bound is based on the notion of $\Wcal$-margin, which is defined for every example $(\xbf,y)\sim \Dcal$ by: $y H_{\Wcal}(\xbf)$, and models the confidence on its label.
Before expounding the $C$-Bound, we introduce the following notations respectively for the first moment ${\cal M}_\Qcal^\Dcal$  and for the second moment ${\cal M}_{\Qcal^2}^\Dcal$ of the \mbox{$\Qcal$-margin},
\begin{align}
\nonumber {\cal M}_\Qcal^\Dcal &= \esp{(\xbf,y)\sim \Dcal} y H_{\Wcal}(\xbf) =\esp{(\xbf,y)\sim \Dcal} \sum_{i=1}^{n} y q_ih_i(\xbf),\\[-1mm]
 {\cal M}_{\Qcal^2}^\Dcal &= \esp{(\xbf,y)\sim\Dcal} \left(y H_{\Wcal}(\xbf)\right)^2 \label{eq:secondmoment} = \esp{(\xbf,y)\sim\Dcal} \sum_{i=1}^{n}\sum_{i'=1}^{n} q_iq_{i'}h_i(\xbf)h_{i'}(\xbf).
\end{align}
By definition, $\BQ$ correctly classifies an example $\xbf$ if the $\posterior$-margin is strictly positive. Thus, under the convention that if $y \espdevant{h\sim \posterior} h(\xbf) = 0$, then $\BQ$ errs on $(\xbf,y)$, we have: 
\begin{align*}
\forall D \mbox{ over }X\times Y,\ R_\Dcal(\BQ)  & = \underset{(\xbf,y)\sim \Dcal}{\mathbf{Pr}} \Big( y H_{\posterior}(\xbf) \leq 0 \Big).
\end{align*}
Knowing this, the authors of \cite{Lacasse07,MinCQ} have proven the following $C$-bound over $\RP(\BQ)$ by making use of the Cantelli-Chebitchev inequality.
\begin{theorem}[\small The C-bound]
\label{theo:C-bound}
Given ${\cal H}$ a class of $n$ functions, for any weights $\{q_i\}_{i=1}^{n}$, and any distribution $\Dcal$ over $X\!\times\! Y$, if $\espdevant{(\xbf,y)\sim {\cal D}}H_{\Wcal}(\xbf)\! >\! 0$ then $R_\Dcal(B_\Qcal)\! \leq\!  C_\Qcal^\Dcal$ where,
\begin{align*}
C_\Qcal^\Dcal&= \frac{\displaystyle \Var_{(\xbf,y)\sim \Dcal} (y  H_{\Wcal}(\xbf))}{\displaystyle \espdevant{(\xbf,y)\sim \Dcal}(y H_{\Wcal}(\xbf))^2}  =  1 - \frac{( {\cal M}_\Qcal^\Dcal )^2  }{ {\cal M}_{\Qcal^2}^\Dcal}.
\end{align*}
\end{theorem}
In the supervised binary classification setting, \cite{MinCQ} have then proposed to minimize the empirical counterpart of the $C$-bound for learning a good majority vote over $\Hcal$, justified by an elegant PAC-Bayesian generalization bound. 
Following this principle the authors have derived the following  quadratic program called MinCq.
{\begin{align}
 &\label{eq:objective}\argmin_{\Qbf}\ \ \displaystyle \Qbf_S^t \Mbf_S \Qbf - \Abf_S^t \Qbf,\\
& \label{eq:mincq_constraint1} \textrm{s.t.}\ \  \displaystyle \mbf_S^t \Qbf =  \frac{\mu}{2}+ \frac{1}{nm} \sum_{j=1}^m \sum_{i=1}^{n} y_jh_i(\xbf_j), \\
& \label{eq:mincq_constraint2}  \textrm{and}\ \  \forall i\in \{1,\dots,n\},\ \displaystyle0\leq q_i'\leq\tfrac{1}{n},\\
 \label{eq:mincq} \tag{\mbox{$MinCq$}}
\end{align}}\\[-8mm]
where ${}^t$ is the transposed function, $\Qbf\! =\! (q_1',\dots,q_n')^t$ is the vector of the  first $n$  weights $q_i$, $\Mbf_S$ is the $n\!\times\! n$ matrix formed by $\frac{1}{m}\sum_{j=1}^m h_i(\xbf_j)h_{i'}(\xbf_j)$ for $(i,i')$ in $\{1,\dots,n\}^2$, 
$\Abf_S = \Big(\frac{1}{nm} \sum_{i=1}^n \sum_{j=1}^m h_1(\xbf_j)h_{i}(\xbf_j),\dots, \frac{1}{nm} \sum_{i=1}^n \sum_{j=1}^m h_n(\xbf_j)h_{i}(\xbf_j)\Big)^t$, and,\\ $\mbf_S =\Big(\frac{1}{m}\sum_{j=1}^m y_jh_1(\xbf_j),\dots, \frac{1}{m}\sum_{j=1}^m y_jh_n(\xbf_j)\Big)^t$.\\[2mm]
Finally, the  majority vote learned by MinCq is $\displaystyle B_{\Wcal}(\xbf)\!=\!\sign[H_{\Wcal}(\xbf)],$ with,\\
 \centerline{$\displaystyle H_{\Wcal}(\xbf)= \sum_{i=1}^{n}\,  \underbrace{\left(2q_i'\! -\!\tfrac{1}{n}\right)}_{\mbox{\small $q_i$}}\, h_i(\xbf).$}
Concretely, MinCq minimizes the denominator of the $C$-bound (Eq.~\eqref{eq:objective}), given a fixed numerator, {\it i.e.} a fixed $\Qcal$-margin (Eq.~\eqref{eq:mincq_constraint1}), under a particular regularization (Eq.~\eqref{eq:mincq_constraint2})\footnote{For more technical details on MinCq please see \cite{MinCQ}.}. Note that, MinCq has showed good performances for binary classification.

\section{A New Framework for Classifier Late Fusion}
\label{sec:mincqpw}
MinCq stands in the particular context of machine learning binary classification. 
In this section, we propose to extend it for designing a new framework for multimedia late fusion.
We actually consider two extensions for dealing with ranking, one with pairwise preferences  and a second based on a relaxation of these pairwise preferences to lighten the process. 
First of all, we discuss in the next section the usefulness of MinCq in the context of multimedia late fusion.

\subsection{Justification of MinCq as a Classifier Late Fusion Algorithm}

It is well known that diversity is a key element in the success of classifier combination \cite{AtreyHEK10,Kuncheva,Dietterich00,Fakeri-TabriziAG13}.
From a multimedia indexing standpoint, fuzing diverse voters is thus necessary for leading to good performances. 
We quickly justify that this is exactly what  MinCq does by favoring majority votes with maximally uncorrelated voters.

In the literature, a general definition of diversity does not exist.
However, there are  popular diversity metrics based on pairwise difference on every pair of individual classifiers, such as $Q$-statistics, correlation coefficient, disagreement measure, {\it etc.} \cite{Kuncheva,LeonardLZTCD11}
We consider the following diversity measure assessing the disagreement between the predictions of a pair of voters according to the distribution $\Dcal$,\\[1mm]
\centerline{$\dive{\Dcal}{}(h_i,h_{i'}) = \esp{(\xbf,y)\sim\Dcal} h_i(\xbf)h_{i'}(\xbf).$}\\
We then can rewrite the second moment of the $\Wcal$-margin (see Eq.\eqref{eq:secondmoment}),\\[-6mm]
\begin{align}
\label{eq:divsecond}
 {\cal M}_{\Qcal^2}^\Dcal = \mbox{\small $\displaystyle \sum_{i=1}^{n} \sum_{i'=1}^{n} q_iq_{i'}$} \dive{\Dcal}{}(h_i,h_{i'}).\\[-7mm]\nonumber
\end{align}
 The first objective of MinCq is to reduce this second moment, implying a direct optimization of  Eq.~\eqref{eq:divsecond}.
This implies  a maximization of the diversity between voters: MinCq favors maximally uncorrelated voters and  appears   to be a natural way for late fusion to combine the predictions of classifiers separately trained from various modalities.

\subsection{MinCq for Ranking} 
In many applications, such as information retrieval, it is well known that ranking documents is a key point to help users browsing results. The traditional measures to evaluate the ranking ability of algorithms are related to precision and recall. Since a low-error vote is not necessarily a good ranker, we propose in this section an adaptation of MinCq to allow optimization of the Mean Averaged Precision (MAP) measure.

Concretely, given a training sample of size $2m$ we split it randomly into two subsets $S'$ and $S\!=\!\{(\xbf_j,y_j)\}_{j=1}^{m}$ of the same size.
Let $n$ be the number of modalities.
For each  modality $i$, we train a classifier $h_i$ from $S'$. 
Let $\Hcal\!=\!\{h_1,\dots,h_n\}$ be the set of the $n$ associated prediction functions and their opposite.
Now at this step, the fusion is achieved by MinCq: We learn from $S$ the weighted majority vote over $\Hcal$ with the lowest risk.\\
We now recall the definition of the MAP measured on $S$ for a given real-valued function $h$.
Let $S^+\!\!=\!\{(\xbf_j,y_j)\! :\! (\xbf_j,y_j)\!\in\! S \wedge y_j\!=\!1\}\!=\!\{(\xbf_{j^+},1)\}_{j^+=1}^{m^+}$ be the set of the $m^+$ positive examples from $S$  and $S^-\!\!=\!\{(\xbf_j,y_j)\! :\! (\xbf_j,y_j)\!\in\! S \wedge  y_j\!=\!-1\}\!=\!\{(\xbf_{j^-},-1)\}_{j^-=1}^{m^-}$ the set of the $m^-$ negative examples from $S$ ($m^+\!\!+\!m^-\!\!=\!m$).
For evaluating the MAP, one ranks the examples in descending order of the scores.
The MAP of $h$ over $S$ is,\\[1mm]
\centerline{$\displaystyle MAP_S(h) = ${\small$\displaystyle \frac{1}{|m^+|} \sum_{j:y_j=1}$}$ Prec@j,$}\\
where $Prec@ j$ is the percentage of positive examples in the top $j$.
The intuition is that we prefer positive examples with a score higher than negative ones.\\

\noindent{\bf MinCq  with Pairwise Preference.} 
To achieve this goal, we propose to make use of {\it pairwise preferences} \cite{Preference} on pairs of positive-negative instances.
Indeed, pairwise methods are known to be a good compromise between accuracy and more complex performance measure like MAP.
Especially, the notion of order-preserving pairwise loss was introduced in \cite{Zhang2004} in the context of multiclass classification.
Following this idea, \cite{YueFRJ07} have proposed a SVM-based method with a hinge-loss relaxation of a MAP-loss.
In our specific case of MinCq for late fusion, we design an order-preserving pairwise loss for correctly ranking the positive examples.
For each pair  $(\xbf_{j^+}\!,\xbf_{j^-}\!)\!\in\! S^+\!\times\! S^-$, we want,
$$H_{\Wcal}(\xbf_{j^+})\! >\! H_{\Wcal}(\xbf_{j^-}) \Leftrightarrow H_{\Wcal}(\xbf_{j^-})\! -\! H_{\Wcal}(\xbf_{j^+})\! <\! 0.$$
This can be forced by minimizing (according to the weights) the following hinge-loss relaxation of the previous equation (where $[a]_+\!=\!\max(a,0)$ is the hinge-loss),\\[-7mm]
{\small
\begin{align}
\label{eq:PWsimp}
\frac{1}{m^+m^-}  \sum_{j^+=1}^{m^+}  \sum_{j^-=1}^{m^-}  \Big[\sum_{i=1}^n  \underbrace{\displaystyle \left( 2q_i - \tfrac{1}{n} \right)   \left(h_i(\xbf_{j^-})  -  h_i(\xbf_{j^+}) \right)}_{H_{\Wcal}(\xbf_{j^-})  -  H_{\Wcal}(\xbf_{j^+})} \Big]_+.
\end{align}}\\[-5mm]%
To deal with the hinge-loss of \eqref{eq:PWsimp}, we consider  $m^+\!\times\!m^-$ additional {\it slack variables $\boldsymbol{\xi}_{S^+\x S^-}\! =\! (\xi_{\mbox{\tiny$j^+j^-$}})_{1\leq j^+ \leq m^+,1\leq j^-\leq m^-} $} weighted by a parameter $\beta\! >\! 0 $. 
We make a little abuse of notation to highlight the difference with \eqref{eq:mincq}: Since $\boldsymbol{\xi}_{S^+\x S^-}$ appear only in the linear term, we obtain the following quadratic program \eqref{eq:mincqpw},
{\small \begin{align}
\nonumber &\argmin_{\Qbf,\boldsymbol{\xi_{S^+\x S^-}}}\ \ \displaystyle \Qbf_S^t \Mbf_S \Qbf - \Abf_S^t \Qbf +  \beta\ {\bf Id}^t\boldsymbol{\xi}_{S^+\x S^-},\\[-1mm]
\nonumber & \textrm{s.t.}\ \  \displaystyle \mbf_S^t \Qbf =  \frac{\mu}{2}+ \frac{1}{nm} \sum_{j=1}^m \sum_{i=1}^{n} y_jh_i(\xbf_j),\\[-4mm]
\nonumber & \mbox{\scriptsize$\forall (j^+\!,j^-)\!\in\!\{1,..,m^+\!\}\!\times\!\{1,..,m^-\!\},$}\, \xi_{\mbox{\tiny$j^+j^-$}}\! \geq\! 0,\, \xi_{\mbox{\tiny$j^+j^-$}}\!\! \geq\!\! \frac{1}{m^+m^-}\!\sum_{i=1}^n\!  \left( 2q_i'\! -\! \tfrac{1}{n} \right)\!\!   \left(h_i(\xbf_{j^-})\!  - \! h_i(\xbf_{j^+})\! \right)\!,\\[-1mm]
&  \textrm{and}\ \  \forall i\in \{1,\dots,n\},\ \displaystyle0\leq q_i'\leq\tfrac{1}{n},  \label{eq:mincqpw} \tag{\mbox{$MinCq_{PW}$}}
\end{align}}%
where ${\bf Id} = \left(1,\dots,1\right)$  of size $(m^{+}\!\times\!m^{-})$. 
However, one drawback of this method is the incorporation of a quadratic number of additive variables ($m^{+}\!\times\!m^{-}$) which makes the problem harder to solve. To overcome this problem, we relax this approach as follows.\\

\noindent\textbf{MinCq with Average Pairwise Preference.} 
 We  relax the constraints by considering the average score over the negative examples: we force the positive ones to be higher than the average negative scores.
This leads us to the following alternative problem \eqref{eq:mincqpwav} with only $m^+$ additional variables.
{\small \begin{align}
\nonumber &\argmin_{\Qbf,\boldsymbol{\xi_{S^+}}}\ \ \displaystyle \Qbf_S^t \Mbf_S \Qbf - \Abf_S^t \Qbf +  \beta\ {\bf Id}^t\boldsymbol{\xi}_{S^+},\\[-1mm]
\nonumber& \textrm{s.t.}\ \  \displaystyle \mbf_S^t \Qbf =  \frac{\mu}{2}+ \frac{1}{nm} \sum_{j=1}^m \sum_{i=1}^{n} y_jh_i(\xbf_j),\\[-4mm]
\nonumber &\quad \forall j^+\!\!\in\!\{1,\dots,m^+\!\},\ \xi_{j^+}\! \geq\! 0,\ \displaystyle  \xi_{j^+}\! \geq\! \frac{1}{m^+m^-}\! \sum_{j^-=1}^{m^-} \! \sum_{i=1}^n \!  \left( 2q_i'\! -\! \tfrac{1}{n} \right) \!  \left(h_i(\xbf_{j^-})\!  -\!  h_i(\xbf_{j^+}) \right),\\[-2mm]
&  \textrm{and}\ \  \forall i\in \{1,\dots,n\},\ \displaystyle0\leq q_i'\leq\tfrac{1}{n},  \label{eq:mincqpwav} \tag{\mbox{$MinCq_{PWav}$}}
\end{align}}%
where ${\bf Id} = \left(1,\dots,1\right)$  of size $m^{+}$.

Note that the two approaches stand in the original framework of  MinCq. In fact, we  regularize the search of the weights for majority vote leading to an higher MAP. 
To conclude, our extension of MinCq aims at favoring $\Wcal$-majority vote implying a good trade-off between classifiers maximally uncorrelated and leading to a relevant ranking.

\renewcommand\tabcolsep{5pt} 
\begin{table*}[ht]
\centering
\scriptsize
\begin{tabular}{|c||c|c|c|c|c|c|c|}
\hline
concept &   $MinCq_{PWav}$  &	 $MinCq_{PW}$  &	 $MinCq$  &	 $\Sigma$  &	 $\Sigma_{MAP}$  &	 $best$  &	 $h_{best}$ \\			
\hline			
\hline																	
aeroplane  &	 $ 0.487	$  &	 $ 0.486	$  &	 $ \mathbf{0.526}	$  &	 $ 0.460	$  &	 $ 0.241	$  &	 $ 0.287	$  &	 $ 0.382$ \\										
\hline	
bicycle	 &	 $ 0.195	$  &	 $ 0.204	$  &	 $ \boldsymbol{ 0.221}	$  &	 $ 0.077	$  &	 $ 0.086	$  &	 $ 0.051	$  &	 $ 0.121$ \\	
\hline	
bird	& 		 $ 0.169	$  &	 $ 0.137	$  &	 $\boldsymbol{ 0.204}	$  &	 $ 0.110	$  &	 $ 0.093	$  &	 $ 0.113	$  &	 $ 0.123$ \\							
\hline	
boat	&  $ 0.159	$  &	 $ 0.154	$  &	 $ 0.159	$  &	 $ 0.206	$  &	 $ 0.132	$  &	 $ 0.079	$  &	 $\boldsymbol{ 0.258}$ \\	
\hline	
bottle	& 	 $ 0.112	$  &	 $ \mathbf{0.126}	$  &	 $0.118	$  &	 $ 0.023	$  &	 $ 0.025	$  &	 $ 0.017	$  &	 $ 0.066$ \\	
\hline
bus	& 	 ${ 0.167}	$  &	 $ 0.166	$  &	 $ \boldsymbol{0.168}	$  &	 $ 0.161	$  &	 $ 0.098	$  &	 $ 0.089	$  &	 $ 0.116$ \\
\hline
car	& 		 $\boldsymbol{ 0.521}	$  &	 $ 0.465	$  &	 $ 0.495	$  &	 $ 0.227	$  &	 $ 0.161	$  &	 $ 0.208	$  &	 $ 0.214$ \\	
\hline			
cat	& 	 $\boldsymbol{ 0.230}	$  &	 $ 0.219	$  &	 $ 0.220	$  &	 $ 0.074	$  &	 $ 0.075	$  &	 $ 0.065	$  &	 $ 0.116$ \\		
\hline	
chair	 &	 $ \boldsymbol{ 0.257}	$  &	 $ 0.193	$  &	 $ 0.230	$  &	 $ 0.242	$  &	 $ 0.129	$  &	 $ 0.178	$  &	 $ 0.227$ \\	
\hline			
cow	 &	 $ 0.102	$  &	 $ 0.101	$  &	 $\boldsymbol{ 0.118}	$  &	 $ 0.078	$  &	 $ 0.068	$  &	 $ 0.06	$  &	 $ 0.101$ \\			
\hline	
diningtable  &	 $ 0.118	$  &	 $ 0.131	$  &	 $ 0.149	$  &	 $\boldsymbol{ 0.153}	$  &	 $ 0.091	$  &	 $ 0.093	$  &	 $ 0.124$ \\
\hline	
dog	& 	 $ \boldsymbol{ 0.260}	$  &	 $ 0.259	$  &	 $ 0.253	$  &	 $ 0.004	$  &	 $ 0.064	$  &	 $ 0.028	$  &	 $ 0.126$ \\
\hline
horse	 &	 $ 0.301	$  &	 $ 0.259	$  &	 $ 0.303	$  &	 $\boldsymbol{ 0.364}	$  &	 $ 0.195	$  &	 $ 0.141	$  &	 $ 0.221$ \\
\hline	
motorbike  &	 $ 0.141	$  &	 $ 0.113	$  &	 $ 0.162	$  &	 $ \boldsymbol{ 0.193}$  &	 $ 0.115	$  &	 $ 0.076	$  &	 $ 0.130$ \\
\hline
person	& 	 $ \boldsymbol{ 0.624}	$  &	 $ 0.617	$  &	 $ 0.604	$  &	 $ 0.001	$  &	 $ 0.053	$  &	 $ 0.037	$  &	 $ 0.246$ \\	
\hline	
pottedplant &	 $\boldsymbol{ 0.067}	$  &	 $ 0.061	$  &	 $ 0.061	$  &	 $ 0.057	$  &	 $ 0.04	$  &	 $ 0.046	$  &	 $ 0.073$ \\	
\hline			
sheep	 &		 $ 0.067	$  &	 ${0.096}	$  &	 $ 0.0695	$  &	 $\boldsymbol{0.128}	$  &	 $ 0.062	$  &	 $ 0.064	$  &	 $ 0.083$ \\	
\hline		
sofa	 &		 $ 0.204	$  &	 $ \boldsymbol{ 0.208}	$  &	 $ 0.201	$  &	 $ 0.137	$  &	 $ 0.087	$  &	 $ 0.108	$  &	 $ 0.147$ \\	
\hline
train	 & 	 $ 0.331	$  &	 $ 0.332	$  &	 $ \boldsymbol{ 0.335}	$  &	 $ 0.314	$  &	 $ 0.164	$  &	 $ 0.197	$  &	 $ 0.248$ \\							
\hline			
tvmonitor  &	 $\mathbf{0.281}	$  &	 $ \mathbf{0.281}	$  &	 $ 0.256	$  &	 $ 0.015	$  &	 $ 0.102	$  &	 $ 0.069	$  &	 $ 0.171$ \\										
\hline	
\hline	
Average	 &	 $ 0.240	$  &	 $ 0.231	$  &	 $ \boldsymbol{  0.243}	$  &	 $ 0.151	$  &	 $ 0.104	$  &	 $ 0.100	$  &	 $ 0.165$ \\	
\hline	
\end{tabular}
\caption{MAP obtained on the PascalVOC'07 test sample.}
\label{tab:res1}
\end{table*}

\renewcommand\tabcolsep{5.5pt} 

\begin{table*}[ht]
\centering
\scriptsize
\begin{tabular}{|c||c|c|c|}
\hline
concept &	 $MinCq_{PWav}^{rbf}$  &	 $MinCq^{rbf}$  &	 SVM$^{rbf}$ \\			
\hline			
\hline																	
aeroplane  &	 $\boldsymbol{ 0.513}	$  &	 $ \boldsymbol{ 0.513}	$  &	 $ 0.497	$  \\										
\hline	
bicycle	 &	 $ \mathbf{0.273} $  &	 $ 0.219	$  &	 $ 0.232	$  \\ 
\hline	
bird	& 	 $ \boldsymbol{0.266}	$  &	 ${0.264}	$  &	 $ 0.196	$  \\				
\hline	
boat	& 	 $ \mathbf{0.267}	$  &	 $ 0.242	$  &	 $ 0.240	$   \\	
\hline	
bottle	& 	 $ \boldsymbol{ 0.103}	$  &	 $ 0.099	$  &	 $ 0.042	$   \\	
\hline
bus	& 	 $\mathbf{0.261}	$  &	 $ \mathbf{0.261}	$  &	 $ 0.212	$ \\
\hline
car	& 	 $ \mathbf{0.530}	$  &	 $ \mathbf{0.530}	$  &	 $ 0.399	$  \\	
\hline			
cat	& 	 $ \mathbf{ 0.253}	$  &	 $ 0.245	$  &	 $ 0.160	$   \\		
\hline	
chair	 &	 $\mathbf{0.397}	$  &	 $\mathbf{ 0.397}	$  &	 $ 0.312	$  \\	
\hline	
cow	 &	 $ 0.158	$  &	 $\mathbf{0.177}	$  &	 $ 0.117	$   \\			
\hline	
diningtable	 &	 $\mathbf{0.263}	$  &	 $ 0.227	$  &	 $ 0.245	$   \\
\hline	
dog	& 	 $\mathbf{ 0.261}	$  &	 $ 0.179	$  &	 $ 0.152	$   \\
\hline
horse	 &	 $\mathbf{ 0.495}	$  &	 $ 0.450	$  &	 $ 0.437	$   \\
\hline	
motorbike  &	 $\mathbf{ 0.295}	$  &	 $ 0.284	$  &	 $ 0.207	$   \\
\hline
person	& 	 $ \mathbf{0.630}	$  &	 $ 0.614	$  &	 $ 0.237	$   \\	
\hline	
pottedplant	 &	 $ 0.102	$  &	 $\mathbf{ 0.116}	$  &	 $ 0.065	$  \\	
\hline			
sheep	 &	 $ \mathbf{0.184}	$  &	 $ 0.175	$  &	 $ 0.144	$   \\	
\hline		
sofa	 &	 $\mathbf{0.246}	$  &	 $ 0.211	$  &	 $ 0.162	$   \\	
\hline
train	 &	 $ \mathbf{0.399}	$  &	 $ 0.385	$  &	 $ 0.397	$   \\							
\hline			
tvmonitor  &$\boldsymbol{ 0.272}	$  &	 $ 0.257	$  &	 $ 0.230	$   \\										
\hline	
\hline	
Average	 &$ \mathbf{0.301}	$  &	 $ 0.292	$  &	 $ 0.234$  \\	
\hline	
\end{tabular}
\caption{MAP obtained on the PascalVOC'07 test sample with a RBF kernel layer.}
\label{tab:res2}
\end{table*}

\section{Experiments on PascalVOC'07 benchmark}
\label{sec:expe}

\noindent\textbf{Protocol.} In this section, we show empirically the usefulness of late fusion MinCq-based methods with stacking.
We experiment these approaches on the PascalVOC'07 benchmark \cite{voc2007}, where the objective is to perform the classification for 20 concepts. The corpus is constituted of $10,000$ images split into train, validation and test sets. 
For most of concepts, the ratio between positive and negative examples is less than $10\%$, which leads to unbalanced dataset and requires to carefully train each classifier. For simplicity reasons, we generate a training set constituted of all the training positive examples and negative examples independently drawn such that the positive ratio is $1/3$. We keep the original test set. Indeed, our objective is not to provide the best results on this benchmark but rather to evaluate if the MinCq-based methods could be helpful for the late fusion step in multimedia indexing.
We consider $9$ different visual features, that are SIFT, Local Binary Pattern (LBP), Percepts, $2$ Histograms Of Gradient (HOG), $2$ Local Color Histograms (LCH) and $2$ Color Moments (CM):\\[-5mm]
\begin{itemize}
\item[\textbullet] LCH are 
$3 \!\times\! 3 \!\times\! 3$ histogram on a grid of $8\!\times\! 6$ or $4 \!\times\! 3$ blocs. Color Moments are represented by the two first moments on a grid of $8 \!\times\! 6$ or $4 \!\times\! 3$ blocs.
\item[\textbullet]HOG is computed on a grid of $4 \!\times\! 3$ blocs. Each bin is defined as the sum of the magnitude gradients from 50 orientations. Thus, overall EDH feature has 600 dimensions. HOG feature is known to be invariant to scale and translation.
\item[\textbullet]LBP is computed on grid of $2 \!\times\! 2$ blocs, leading to a $1,024$ dimensional vector. The LBP operator labels the pixels of an image by thresholding the $3 \!\times\! 3$-neighborhood of each pixel with the center value and considering the result as a decimal number. LBP is known to be invariant to any monotonic change in gray level.
\item[\textbullet]Percept features are similar to SIFT codebook where visual words are related to semantic classes at local level. There are 15 semantic classes such as 'sky', 'skin', 'greenery', 'rock', etc. We also considered SIFT features from a dense grid, then map it on a codebook of 1000 visual words generated with Kmeans. \\[-5mm]
\end{itemize}
\indent We train a SVM-classifier for each feature with the LibSVM library \cite{libsvm} and a RBF kernel with parameters tuned by cross-validation.  The set $\Hcal$ is then constituted by the $9$ score functions associated with the SVM-classifiers.

\indent In a first series of experiments, the set of voters $\Hcal$ is constituted by the  $9$ SVM-classifiers.
We compare our $3$ MinCq-based methods 
to the following $4$ baselines:\\[-3mm]
\begin{itemize}
\item[\textbullet]The best classifier of $\Hcal$:\hfill $ \displaystyle h_{best} = \argmax_{h_i\in\Hcal} MAP_{S}(h_i).$\\[-1mm]
\item[\textbullet]The one with the highest confidence: \hfill$ \displaystyle best(\xbf) = \argmax_{h_i\in\Hcal} |h_i(\xbf)|.$\\[-1mm]
\item[\textbullet]The sum of the classifiers (unweighted vote): \hfill$ \Sigma(\xbf) =\sum_{h_i\in\Hcal} h_i(\xbf).$\\[-1mm]
\item[\textbullet]The MAP-weighted vote: \hfill$\displaystyle \Sigma_{MAP}(\xbf) =${\scriptsize$\displaystyle \sum_{h_i\in\Hcal} \frac{MAP_{S}(h_i)}{\sum_{h_{i'}\in\Hcal}MAP_{S}(h_{i'})}$}$h_i(\xbf).$\\[-3mm]
\end{itemize}

In a second series, we propose to introduce non-linear information with a RBF kernel layer for increasing the diversity over the set $\Hcal$. We consider a larger $\Hcal$ as follows.
Each example is  represented by the vector of its scores with the 9 SVM-classifiers and $\Hcal$ is now  the set of kernels over the sample $S$: Each  $\xbf\!\in\!S$ is seen as a voter $k(\cdot,\xbf)$.
We  compare this approach to classical stacking with SVM.

Finally, for tuning the hyperparameters we use a $5$-folds cross-validation process, where 
instead of selecting the parameters leading to the lowest risk, we select the ones leading to the best MAP. 
 The MAP-performances are reported on Tab. \ref{tab:res1} for the first series and on Tab. \ref{tab:res2} for the second series.

\noindent\textbf{Results.} Firstly, the performance of $\Sigma_{MAP}$ fusion is lower than  $\Sigma$, which means that the performance of single classifiers is not correlated linearly with its importance on the fusion step. 
On Tab. \ref{tab:res1}, for the first experiments, we clearly see that the linear MinCq-based algorithms outperform on average the linear baselines. MinCq-based method produces the highest MAP for $16$ out of $20$ concepts. Using a Student paired t-test, this result is statistically confirmed with a p-value $<0.001$ in comparison with $\Sigma_{MAP}$, $best$ and $h_{best}$.
In comparison of $\Sigma$, the  p-values respectively associated with  \eqref{eq:mincqpwav}, \eqref{eq:mincqpw} and \eqref{eq:mincqpw} are $0.0139$, $0.0232$ and $0.0088$.
We can remark that \eqref{eq:mincqpw} implies lower performances than its relaxation \eqref{eq:mincqpwav}. A Student test leads to a p-value of $0.223$, which statistically means that the two approaches produce similar results.
Thus, when our objective is to rank the positive examples before the negative examples, the average constraints appear to be a good solution.
However, we note that the order-preserving hinge-loss is not really helpful: The classical \eqref{eq:mincq} shows the best MAP results (with a p-value of $0.2574$). Indeed, the trade-off between diversity and ranking is difficult to apply here since the $9$ voters are probably not enough expressive. On the one hand, the preference constraints appear hard to satisfy, on the other hand, the voters' diversity do not really vary.\\
The addition of a kernel layer allows us to increase this expressivity. Indeed, Tab. \ref{tab:res2} shows that the MinCq-based methods achieve the highest MAP for every concept in comparison with SVM classifier.
This confirms  that the diversity between voters is well modeled by MinCq algorithm.
 Especially,  $MinCq_{PWav}^{rbf}$ with the averaged pairwise preference is significantly the best: a Student paired test implies a p-value of $0.0003$ when we compare $MinCq_{PWav}^{rbf}$ to  SVM, and the p-value is $0.0038$ when it is compared to $MinCq^{rbf}$. Thus, the the order-preserving loss is a good compromise between improving the MAP and keeping a reasonable computational cost. Note that we do not report the results for \eqref{eq:mincqpw} in this context, because the computational cost is much higher and the performance is lower. The full pairwise version implies too many variables which penalize the resolution of \eqref{eq:mincqpw}.
Finally, it appears that at least one MinCq-based approach is the best for each concept, showing that MinCq methods outperform the other compared methods. Moreover, a Student test implies a p-value $<0.001$ when we compare $MinCq_{PWav}^{rbf}$ to the approaches without kernel layer. $MinCq_{PWav}^{rbf}$ is significantly then the best approach in our experiments.

We  conclude from these experiments that MinCq-based approaches are a good alternative for late classifiers fusion as it takes into account the diversity of the voters.  In the context of multimedia documents retrieval, the diversity of the voters comes from either the variability of input features or by the variability of first layer classifiers. 

\section{Conclusion and Perspectives}
\label{sec:conclu}
In this paper, we proposed  to make use of a well-founded learning quadratic program called MinCq for multimedia late fusion tasks.
MinCq was originally developed for binary classification, aiming at minimizing the error rate of the weighted majority vote by considering the diversity of the voters \cite{MinCQ}. We designed an adaptation of MinCq able to deal with ranking problems by considering pairwise preferences while taking into account the diversity of the models. 
In the context of multimedia indexing, this extension of MinCq  appears naturally appropriate for  combining the predictions of classifiers trained from various modalities in a late classifier fusion setting. 
Our experiments have confirmed that MinCq is a very competitive alternative for classifier fusion in the context of an image indexing task.
Beyond these results, this work gives rise to many interesting remarks, among which the following ones.
Taking advantage of a margin constraint for late classifier fusion  may allow us to prove a new $C$-bound specific to ranking problems, and thus to derive other algorithms for classifier fusion by maximizing the diversity between the classifiers.
This could be done by investigating some theoretical results using the Cantelli-Chebychev's inequality \cite{Cant} as in \cite{MinCQ}. 
Additionally, it might be interesting to study the impact of using other diversity metrics \cite{Kuncheva} on performances for image and video retrieval. Such an analysis would be useful for assessing a trade-off between the quality of the ranking results and the diversity of the inputs for information retrieval. 
Finally, another perspective, directly founded on the general PAC-Bayes theory \cite{Mcallester99b}, could be to take into account a prior belief on the classifiers of $\Hcal$. 
Indeed, general PAC-Bayesian theory allows one to obtain theoretical guarantees on majority votes with respect to the distance between the considered vote and the prior belief measured by the Kullback-Leibler divergence. The idea is then to take into account prior information for learning good majority votes for ranking problems.

\paragraph{\footnotesize{\bf Acknowledgments.}}  {\footnotesize This work was in parts funded by the European Research Council under the European Unions Seventh Framework Programme (FP7/2007-2013)/ERC grant agreement no {308036}. the authors would like to thanks Thomas Peel for useful comments.}

\bibliographystyle{abbrv}
\bibliography{biblio}

\end{document}